\documentclass{pprai}

\usepackage[utf8]{inputenc}
\usepackage[T1]{fontenc}
\usepackage{graphicx}
\usepackage{amsthm}
\usepackage{txfonts}
\usepackage{url}
\usepackage{algpseudocode}
\usepackage{hyperref}

\title{Pedestrian detection with \\high-resolution event camera}
\headtitle{Pedestrian detection with high-resolution event camera}

\author{Piotr Wzorek \href{https://orcid.org/0000-0003-3885-600X}{\includegraphics[width=16pt]{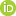}} \and
Tomasz Kryjak \href{https://orcid.org/0000-0001-6798-4444}{\includegraphics[width=16pt]{orcid.png}} }

\headauthor{P. Wzorek, T. Kryjak}
\affiliation{
   Embedded Vision Systems Group, Computer Vision Laboratory\\
   Department of Automatic Control and Robotics\\
   AGH University of Krakow, Poland\\
   \{pwzorek,tomasz.kryjak\}@agh.edu.pl
}


\graphicspath{ {figures/} }

\begin{document}
\maketitle

\begin{abstract}
Despite the dynamic development of computer vision algorithms, the implementation of perception and control systems for autonomous vehicles such as drones and self-driving cars still poses many challenges. 
A video stream captured by traditional cameras is often prone to problems such as motion blur or degraded image quality due to challenging lighting conditions. 
In addition, the frame rate -- typically 30 or 60 frames per second -- can be a limiting factor in certain scenarios.
Event cameras (DVS -- Dynamic Vision Sensor) are a potentially interesting technology to address the above mentioned problems.
In this paper, we compare two methods of processing event data by means of deep learning for the task of pedestrian detection.
We used a representation in the form of video frames, convolutional neural networks and asynchronous sparse convolutional neural networks. 
The results obtained illustrate the potential of event cameras and allow the evaluation of the accuracy and efficiency of the methods used for high-resolution (1280 x 720 pixels) footage.

\end{abstract}

\section{Introduction}

Event cameras are neuromorphic vision sensors inspired by the structure and behaviour of the human eye \cite{gallego2020event}.
They are increasingly being used in computer vision.
They respond to changes in the brightness of the observed scene independently for each pixel.
Each so-called ``event'' is generated when the change in the logarithm of the brightness sensed by a given pixel reaches a certain threshold. A~single event is described by four values:
$e = \{t, x, y, p\}$,
where: $t$ is the timestamp of the event (in microseconds), $x$ and $y$ are the coordinates of the pixel registering the event, and $p$ is its polarity in the form of a value of 1 (positive change in brightness) or -1 (negative change).

The growing popularity of event cameras is the result of their ability to be used in conditions of rapid movement of the object relative to the sensor (temporal resolution of microseconds) and unfavourable lighting conditions (dynamic range of 120 dB, relatively good performance in low light).
The event camera only records changes in the scene.
This means that redundant information is largely eliminated and the number of events generated depends on the dynamics of the scene, camera ergo-motion or lighting changes.
As a result, the power consumption of the sensor is low (in typical situations) and the processing of the most important information can be efficient.

However, there are significant challenges in applying known computer vision algorithms to event data.
These type of data differ significantly from traditional video frames in that they form a sparse spatio-temporal cloud.
Meanwhile, state-of-the-art object detection solutions use traditional video frames as input to deep convolutional neural networks. 

In this work, we have compared different methods for applying deep learning algorithms to event data.
We considered using the representation of event data in the form of frames generated by accumulating them over a defined time window and using as input to a CNN, as well as the use of asynchronous sparse neural networks to optimise energy consumption by reducing the number of operations performed.
We focused on the problem of pedestrian detection as a key issue for different types of autonomous vehicles -- both drones and self-driving cars. The main contribution of this paper is the evaluation of selected methods available in the literature for a high resolution dataset and their comparison in terms of accuracy and efficiency.

The remainder of this paper is organised as follows. Section 2 describes the different methods considered in the literature for applying deep learning with event data to the task of object detection. Section 3 describes the research we have conducted. A summary and considerations for further development plans for the implemented systems conclude the paper.

\section{Object detection with event cameras}

State-of-the-art object detection algorithms use deep convolutional neural networks adapted to process data represented as two- or three-dimensional matrices. 
The unusual nature of event data requires both the use of so-called event data representations and the introduction of modifications to the algorithms used. 
Many approaches to this problem have been proposed in the literature. 

The simplest solution is to accumulate the event data in a matrix, analogous to the frames recorded by classical cameras, and then use deep convolutional neural networks. 
The paper \cite{afshar2020event} proposes a method in which each pixel is assigned the polarity value of the last event recorded. An extension of this idea is the exponentially decaying time surface, where the time of occurrence of an event is also taken into account. Also popular are methods that take into account the frequency of occurrence of an event for a given pixel \cite{chen2018pseudo}, or the so-called leaky surface \cite{cannici2019asynchronous}, where the memory of previous events is retained.

An extension of the idea of using deep convolutional neural networks to process event data is the use of asynchronous sparse convolutional neural networks (ASCNN).  
The authors of \cite{messikommer2020event} exploit the sparsity of event data to reduce the computational complexity and energy consumption of the detection system, using networks in which only the convolution results for changing input values are updated. The use of ASCNNs allows to perform detection efficiently and asynchronously (updating the necessary values for each incoming pixel).


The literature also considers methods based on the fusion of event data and traditional RGB frames \cite{jiang2019mixed}, the use of transformer-based architectures \cite{gehrig2022recurrent}, recurrent architectures \cite{perot2020learning}, Graph Neural Networks \cite{schaefer2022aegnn} or Spiking Neural Networks \cite{cordone2022object}.

\section{The analysed pedestrian detection methods}


For our research, we used the Prophesee 1 Megapixel automotive detection dataset \cite{perot2020learning} containing sequences recorded with a 1280x720 pixel resolution event camera in road conditions with labelled objects such as pedestrians, traffic signs, traffic lights or cars. The dataset was filtered to select only fragments containing pedestrians. 
This yielded more than one million 10ms sequences on which pedestrians were labelled. 
For some experiments, a subset of 100,000 10ms sequences was used to speed up the neural network training process. 


The first method considered was to accumulate event data in time windows of 10ms and use them as input to a deep convolutional neural network.
To maximise the amount of information in the representation, we used a fusion of representations using different characteristics of the data -- polarity, temporal resolution and frequency of occurrence, analogous to \cite{wzorek2022traffic}.
As a detector, we used the YOLOv7 architecture of \cite{wang2022yolov7}, which represents the state of the art in terms of execution time and accuracy. The detector trained on the described dataset using the transfer learning mechanism achieved a detection accuracy of 67.7\%mAP@0.5 (38\%mAP@.5:.95).
The YOLOv7 model is characterised by a value of 104.7 GFLOPs (floating point operations).
We use this metric to evaluate and compare the computational efficiency between multiple models.

\begin{figure}[!t]
    \centering
    \includegraphics[width=0.75\textwidth]{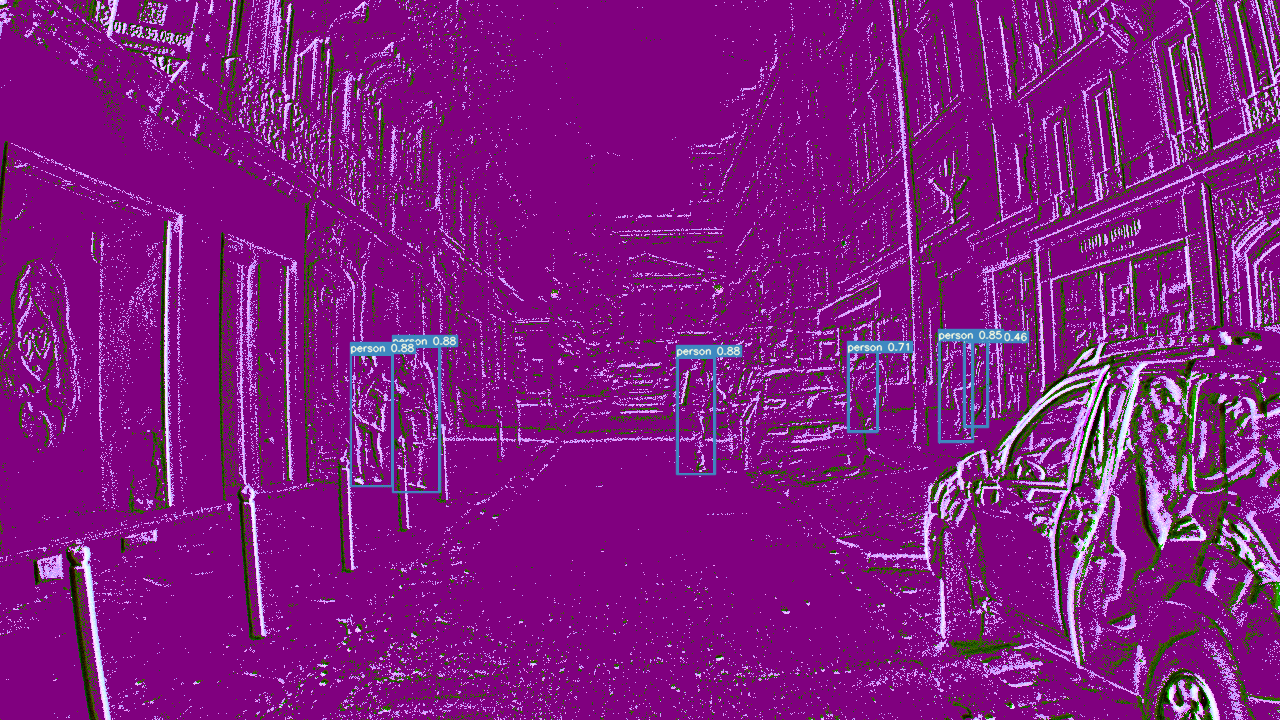}
    \caption{Example of correct detection on the fused representations} 
    \label{fifusion_detect}
\end{figure}

As an alternative to the detection system described above, we also decided to test ASCNNs.
Analogous to \cite{messikommer2020event}, we used the event histogram as event data representation, the VGG-16 network as feature extraction and the YOLO output layer. This model has a reported computational efficiency of 205 MFLOPs, which is a significant reduction compared to the YOLOv7-based detector.

However, we were unable to achieve satisfactory accuracy scores for a ASCNNs after numerous experiments on a reduced data set (100,000 samples). The maximum result was 0.04\%mAP. To increase the network input, we also tried to extend the model by two convolution layers and a max-pooling layer. For the transformed model, we obtained a maximum result of 0.1\%mAP. As a comparison, for the same subset of dataset we obtained 47\%mAP@0.5 (19.8\%mAP@.5:.95) for YOLOv7 model.


\section{Summary}

In this work, we investigated the accuracy and computational efficiency of event camera-based detection systems with deep convolutional network YOLOv7 and ASCNNs.
The accuracy score achieved for the YOLOv7 detector (67.7\% mAP@0.5, 38\%mAP@.5:.95) are satisfactory and comparable to the RED network accuracy (43\%mAP) reported in \cite{perot2020learning}, which is the highest result achieved for this dataset.
The accumulation of data in a 10ms window provides enough information to perform detection with a large YOLOv7 model and a diverse and efficient representation. 
However, the event histogram representations generated in the same time windows and the use of a relatively small network model (VGG16 with YOLO output) do not allow for satisfactory performance results for ASCNNs.


The conducted research makes it possible to plan the direction of further work on detection systems for high-resolution event data, taking into account their accuracy, computational complexity and the possibility of hardware implementation using SoC FPGA or eGPU platforms. 
We plan to further explore ASCNNs architectures to improve the accuracy of detection systems. We will conduct further experiments considering other event data accumulation times (to increase the amount of information) and larger sparse network models. The significant reduction in computational complexity in sparse models is a motivation for further work on this type of network.
Another approach that we also want to test is the use of binary networks. A
relatively simplified event representation should work well with this type of model and allow for improved computational efficiency.
We are also considering various methods to accelerate the mentioned networks using the FPGA platform.
Another type of network that we plan to evaluate and study are spiking neural networks (SNN), which allow direct processing of events. The use of Dynamic Vision Sensors and neuromorphic platforms such as BrainChip or Loihi would enable the task of pedestrian detection to be realised effectively and efficiently. 

\newpage
\section*{Acknowledgements}
The work presented in this paper was supported by the programme ``Excellence initiative – research university'' for the AGH University of Krakow.
We gratefully acknowledge Poland’s high-performance computing infrastructure PLGrid (HPC Centers: ACK Cyfronet AGH) for providing computer facilities and support within computational grant no. PLG/2023/016130.


\bibliography{pprai}
\bibliographystyle{pprai}

\end{document}